\pdfoutput=1

\documentclass[11pt]{article}

\usepackage[preprint]{acl}

\usepackage{times}
\usepackage{latexsym}

\usepackage[T1]{fontenc}

\usepackage[utf8]{inputenc}

\usepackage{microtype}

\usepackage{inconsolata}

\usepackage{graphicx}
\usepackage{booktabs}
\usepackage{multirow}
\usepackage{xcolor}
\usepackage{listings,framed}
\usepackage{lipsum}
\usepackage{cuted}

\title{Decompose, Plan in Parallel, and Merge: A Novel Paradigm for Large Language Models based Planning with Multiple Constraints}

\author{
\textbf{ZhengdongLu}\textsuperscript{1}, 
\textbf{Weikai Lu}\textsuperscript{1}, 
\textbf{Yiling Tao}\textsuperscript{1},
\textbf{Yun Dai}\textsuperscript{1}, 
\\
\textbf{Zixuan Chen}\textsuperscript{1}, 
\textbf{Huiping Zhuang}\textsuperscript{1},
\textbf{Cen Chen}\textsuperscript{1}, 
\textbf{Hao Peng}\textsuperscript{2}, 
\textbf{Ziqian Zeng}\textsuperscript{1}
\\
\textsuperscript{1} South China University of Technology,
\textsuperscript{2} Beihang University
\\
 \small{
   \textbf{Correspondence:} \href{mailto:zqzeng@scut.edu.cn}{zqzeng@scut.edu.cn}
 }
}

\begin{document}
\maketitle
\begin{abstract}
Despite significant advances in Large Language Models (LLMs), planning tasks still present challenges for LLM-based agents.
Existing planning methods face two key limitations: heavy constraints and cascading errors. 
To address these limitations, we propose a novel parallel planning paradigm, which \textbf{D}ecomposes, \textbf{P}lans for subtasks in \textbf{P}arallel, and \textbf{M}erges subplans into a final plan (\textbf{DPPM}). 
Specifically, DPPM decomposes the complex task based on constraints into subtasks, generates the subplan for each subtask in parallel, and merges them into a global plan. 
In addition, our approach incorporates a verification and refinement module, enabling error correction and conflict resolution. 
Experimental results demonstrate that DPPM significantly outperforms existing methods in travel planning tasks. 
\end{abstract}

\section{Introduction} 


Planning represents a fundamental cognitive skill that bridges the gap between reasoning and action \cite{bubeck2024sparks}. 
Large Language Model (LLM) based agents have emerged as ideal candidates for the planning task by combining the reasoning capabilities of LLMs with robust tool manipulation abilities of agents \cite{yao2023react}.  
Despite significant advances in LLMs \cite{touvron2023llama, yang2025qwen3, achiam2023gpt4}, planning presents unique challenges, particularly when tasks involve multiple constraints, require long-horizon thinking, or demand precise coordination across subtasks \cite{yao2023react,shinn2023reflexion}.

\begin{figure}[!t]
  \centering
    \centering
    \includegraphics[width=1\linewidth]{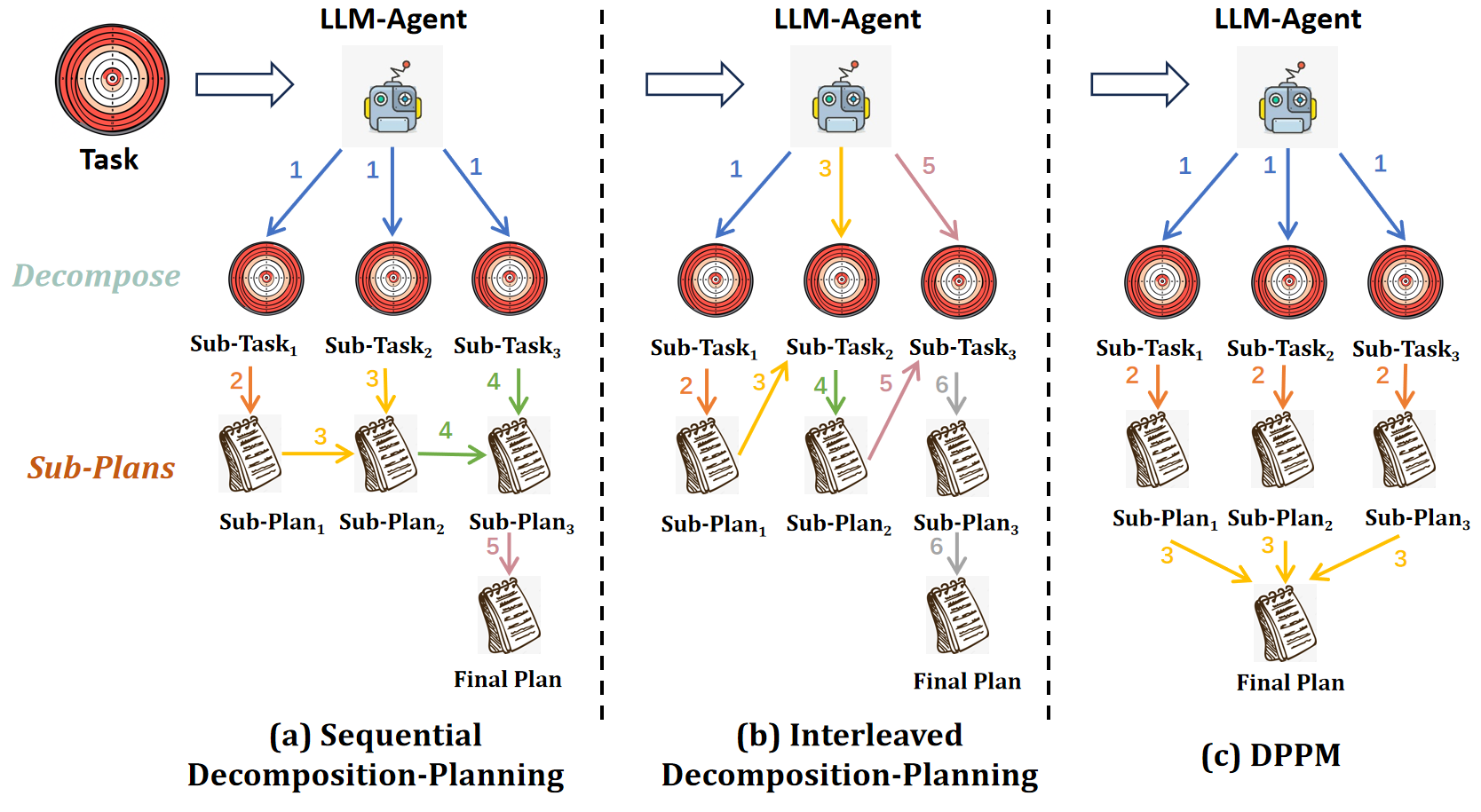}
    \caption{
    Comparison of different types of planning framework, including sequential decomposition-planning, interleaved decomposition-planning, and DPPM.
    Taking travel planning as an example, sequential methods might first decompose a complex task into temporal segments (e.g., daily itineraries) and then generate plans for each day in sequence, with each day's plan building upon previous decisions.
    Interleaved decomposition-planning alternates between decomposition and planning, addressing one day's plan before proceeding to decompose and plan the next day. 
    DPPM first decomposes subtasks by aspects (transportation, accommodation, attraction, and meals), generates corresponding subplans for each aspect, and then merges these subplans into a final plan. 
    }
    \label{fig:decomposition} 
\end{figure}

Existing LLM-based planning methods can be categorized into two types including the \textit{Sequential Decomposition-Planning} and \textit{Interleaved Decomposition-Planning}. 
Sequential methods ~\cite{wang2023plan,shen2023hugginggpt} first break complex tasks into simpler subtasks, then make planning for each subtask sequentially, with each planning step based on the results of the previous one. 
Interleaved methods ~\cite{wei2022chain,zhou2022least,chen2022program,yao2023react,shinn2023reflexion} alternate between decomposition and planning, addressing only one or two subtasks at a time. 
Figure \ref{fig:decomposition} illustrates the core idea of sequential and interleaved methods.

Despite showing promise, both categories struggle with the multi-constraint planning task. 
For example, \citet{gundawar2024robust} revealed that even advanced methods like ReAct~\cite{yao2023react} and Reflexion~\cite{shinn2023reflexion} have achieved a final pass rate less than 1\% in multi-constraint tasks like travel planning. The underlying reasons are two major limitations of existing methods: (1) \textbf{Heavy Constraints.} 
Existing methods do not reasonably allocate constraints when decomposing task, causing some subtasks to handle a disproportionately large number of constraints simultaneously. 
Since current LLMs are not yet capable of effectively managing too many constraints at once, these methods are easy to fail on such subtasks. 
(2) \textbf{Cascading Errors.} The correctness of planning for each subtask depends on the planning results of the previous one, meaning that even minor errors at any step can lead to a failure of the global planning.

To address these limitations, we propose a novel paradigm for multi-constraint planning which \textbf{D}ecomposes, \textbf{P}lans for subtasks in \textbf{P}arallel, and \textbf{M}erges subplans into a final plan (\textbf{DPPM}). 
As shown in Figure \ref{fig:decomposition}, DPPM does not belong to either sequential decomposition-planning or interleaved decomposition-planning. 
DPPM consists of three key stages. 
First, a Constraint-aware Task Decomposition stage aims to break down complex tasks into subtasks, each handling a limited set of relevant constraints. 
Second, a Local Plan Generation stage enables parallel planning by individual LLM agents, each optimizing for its specific subtask without sequential dependencies. 
Finally, an Incremental Merge stage consolidates these local plans into a coherent global solution. 
Additionally, we design a Verification and Refinement module performs constraint checking and error correction during the latter two stages. 
DPPM effectively addresses two critical limitations of existing methods including the heavy constraints problem through our constraint-aware decomposition mechanism, and the cascading error challenge through both parallel subtask planning and verification procedures.

In this paper, we particularly focus on the travel planning, which serves as an ideal testbed for evaluating planning capabilities.
Travel planning mirrors realistic human decision-making processes with numerous interdependent variables including accommodation, transportation, activities, and dining options. 
It naturally incorporates diverse constraint types including temporal restrictions, spatial limitations, budgetary boundaries, and personal preferences that must be simultaneously satisfied \cite{kambhampati2024llmmodulo}.   
These characteristics make travel planning exceptionally challenging for evaluating the advanced planning capabilities of LLM-based agents \cite{gundawar2024modulotravel}.

Extensive experiments conducted on the TravelPlanner~\cite{xie2024travelplanner} dataset demonstrate that, compared to advanced methods such as Direct \cite{xie2024travelplanner}, CoT~\cite{wei2022chain} and LLM-Modulo \cite{kambhampati2024llmmodulo}, which achieved 56.1\%, 59.3\%, and 22.2\% on final pass rate using Qwen2.5-32B-Instruct.
Additionally, on the ChinaTravel \cite{shao2024chinatravel} dataset, our method achieves 42.4\% and 12.8\% higher final pass rates than Direct and LLM-Modulo on hard-level samples, respectively.

The contributions of this paper are summarized as follows,

$\bullet$ We propose a novel paradigm called DPPM for LLM-based  multi-constraint planning. DPPM decomposes complex tasks into subtasks based on constraints, generates plans for each subtask in parallel, and then merges these plans into a coherent solution.


$\bullet$ We introduce a verification and refinement module that corrects errors and resolves conflicts among constraints during the local planing and  merging stage. 

$\bullet$ Experimental results shows that our method outperforms other baselines significantly on various benchmarks and backbones. 
\begin{figure*}[t!]
  \centering
    \centering
    \includegraphics[width=1\linewidth]{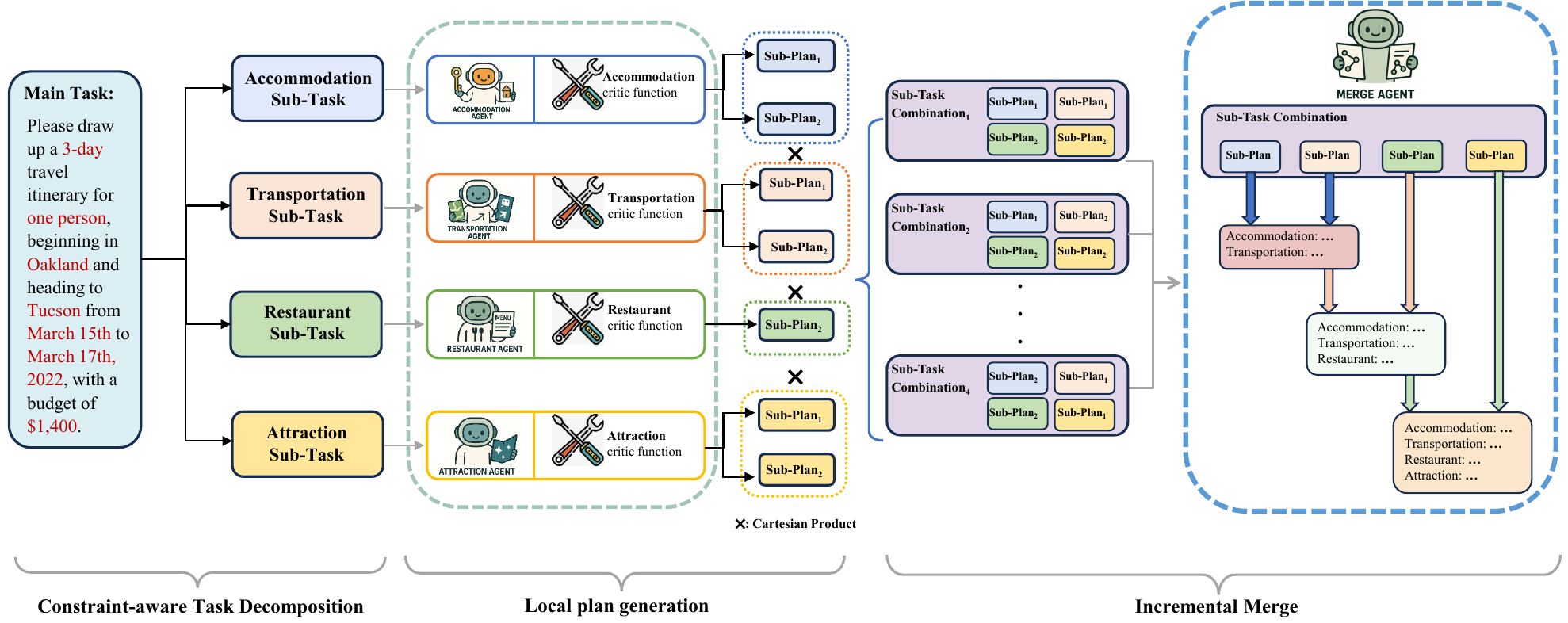}
    \caption{An overview of our proposed DPPM, which consists of three key stages.
    (1) In the Constraint-based Task Decomposition stage, complex planning task will be break down into simpler subtasks. (2) In the Local Plan Generation stage, each local agent produces preliminary planning options for each subtask. (3) In the Incremental Merge stage, subplans will be integrated into a unified global plan.}
    \label{fig:overview} 
\end{figure*}

\section{Related Work} 


Recent advances in Large Language Models (LLMs) have showcased their strong reasoning and planning abilities~\cite{xi2025rise,guo2023indeed}. Consequently, LLM-based agents have become a key research focus, facilitating complex interactions and decision-making across various domains~\cite{alterovitz2016robot,hong2023metagpt,park2023generative,xu2023language}.To improve agent reasoning, researchers have proposed two types of methods: Sequential Methods and Interleaved Methods.

\subsection{Sequential Methods}
Sequential methods typically first break down a complex task into several subtasks, and then plan for each subtask in a fixed, sequential order. For instance, Wang et al.~\cite{wang2023plan} proposed Plan-and-Solve Prompting, which enables a LLM to decompose a complex task into ordered subtasks. Similarly, Shen et al.~\cite{shen2023hugginggpt} introduced HuggingGPT, which decomposes tasks into subtasks and sequentially leverages specialized models on Hugging Face to accomplish them.  Toolformer~\cite{schick2023toolformer} iteratively decomposes complex tasks by dynamically inserting and utilizing API call results. Chameleon~\cite{lu2023chameleon} decomposes tasks into ordered subtasks, each executed via specific functional modules informed by prior context and outputs. These methods uniformly follow a “decompose first, then execute sequentially” paradigm, where each subtask’s output informs the subsequent one. However, this risks error propagation, as early mistakes can accumulate. Additionally, uneven constraint distribution may overload some subtasks, exceeding LLMs’ reasoning capacity and raising failure risk.

\subsection{Interleaved Methods}
Another category is interleaved methods, which alternate between task decomposition and planning in an iterative manner, gradually addressing a limited number of subtasks. Representative works such as Chain-of-Thought Prompting ~\cite{wei2022chain} and Least-to-Most Prompting ~\cite{zhou2022least} guide the model to perform step-by-step reasoning. Tree and graph search strategies~\cite{yao2023tree,besta2024graph} further assist in exploring more optimal decision paths. Furthermore, the self-reflection mechanism proposed by Madaan et al.~\cite{madaan2023self} enables the model to identify and correct errors in reasoning. ReAct~\cite{yao2023react} and Reflexion ~\cite{shinn2023reflexion} combine environmental feedback with multi-turn self-reflection, enhancing the model's adaptability to dynamic tasks and the accuracy of planning. Although interleaved methods effectively mitigate the error accumulation found in sequential approaches, they still face limitations due to the restricted number of subtasks handled at each step. This can lead to imbalanced constraint allocation and a lack of a holistic global perspective, potentially causing conflicts between subtasks and undermining the coherence and executability of the overall plan.

\section{Methodology}

We propose DPPM, a novel framework for LLM-based planning with multiple constraints. 
As shown in Figure~\ref{fig:overview}, DPPM consists of three stage: Constraint-aware Task Decomposition, Local Plan Generation, and Incremental Merge. Specifically, a complex task is first broken down into multiple subtasks according to constraints. Then, multiple local agents generate plans for each subtask. Finally, a merging agent organizes all the subplans into a complete plan. To ensure each agent successfully completes its task, we additionally designed a Verification and Refinement module for error correction and conflict resolution. 




\subsection{Constraint-aware Task Decomposition}
The constraint-aware task decomposition stage aims to break down a complex task that contain numerous constraints into simpler subtasks with fewer constraints, making them easier to plan. 
For travel planning tasks, user queries often involve multiple constraints, covering aspects including transportation, accommodation, attraction, and meals \cite{huang2024understanding}. 
These constraints can be classified into two categories. 
Local constraints apply exclusively to a single aspect (e.g., room type for accommodation, cuisine preferences for meals, or mode of transportation).
Global constraints span multiple aspects (e.g., overall budget limitations or minimum stay requirements) \cite{xie2024travelplanner}. 
Current LLMs struggle when confronted with heavy constraints simultaneously.
Hence, if we divide all constraints into groups according to aspects, then each group contains only a manageable subset of the total constraints. 
Specifically, we divide all constraints into four main groups including transportation, accommodation, attraction, and meals.  
This allows us to decompose the overall task into subtasks, with each subtask handling a manageable constraint subset.
For example, the subtask for transportation is ``please give me a detailed transportation plan based on the user's query requirements.'', and its expected subplan only needs to satisfy constraints that are related to transportation. 
Once the task is divided into subtasks, each subtask is sent to the next stage. 

\subsection{Local Plan Generation}
After the task decomposition is complete, the local plan generation stage is responsible for producing preliminary planning options for each subtask.

To achieve this, we delegates different types of subtasks to specialized local agents.
There are four local agents responsible for transportation, accommodation, attraction, and meals, respectively. 
Each agent comprises an LLM coupled with a dedicated constraint evaluation function. 
Unlike sequential methods \cite{wang2023plan,shen2023hugginggpt} where subplans depend on the previous subplan, our method lets local agents work independently.
Communication between local agents is unnecessary, as each agent focuses solely on solving its assigned subtask, enabling parallelized planning. 
Specifically, we design custom prompt templates for each subtask, which can be found in Appendix \ref{prompt:loacl_plan}. 
Beyond specifying the subtask objectives and output formats, these prompts incorporate in-context learning \cite{brown2020fewshot, dong2022iclsurvey} by providing an example within the context for the LLM to learn from, and require the LLM to generate multiple planning proposals. 

To mitigate the risk of producing subplans that may adversely affect subsequent integration (e.g., exhausting the entire budget on a single subtask), we incorporate a Global Constraint Instruction into the prompt, explicitly acknowledging the existence of other subtasks and instructing the LLM to reserve sufficient planning space for them. 
Furthermore, we employ high-temperature sampling to ensure the model offers a diverse set of options. 
While subplans may initially contain errors or constraint violations, these issues are systematically addressed by our verification and refinement module in Section \ref{sec:verfify}. 
After correction, the subplans are sent to the next stage.

\begin{figure}[!t]
  \centering
    \centering
    \includegraphics[width=1\linewidth]{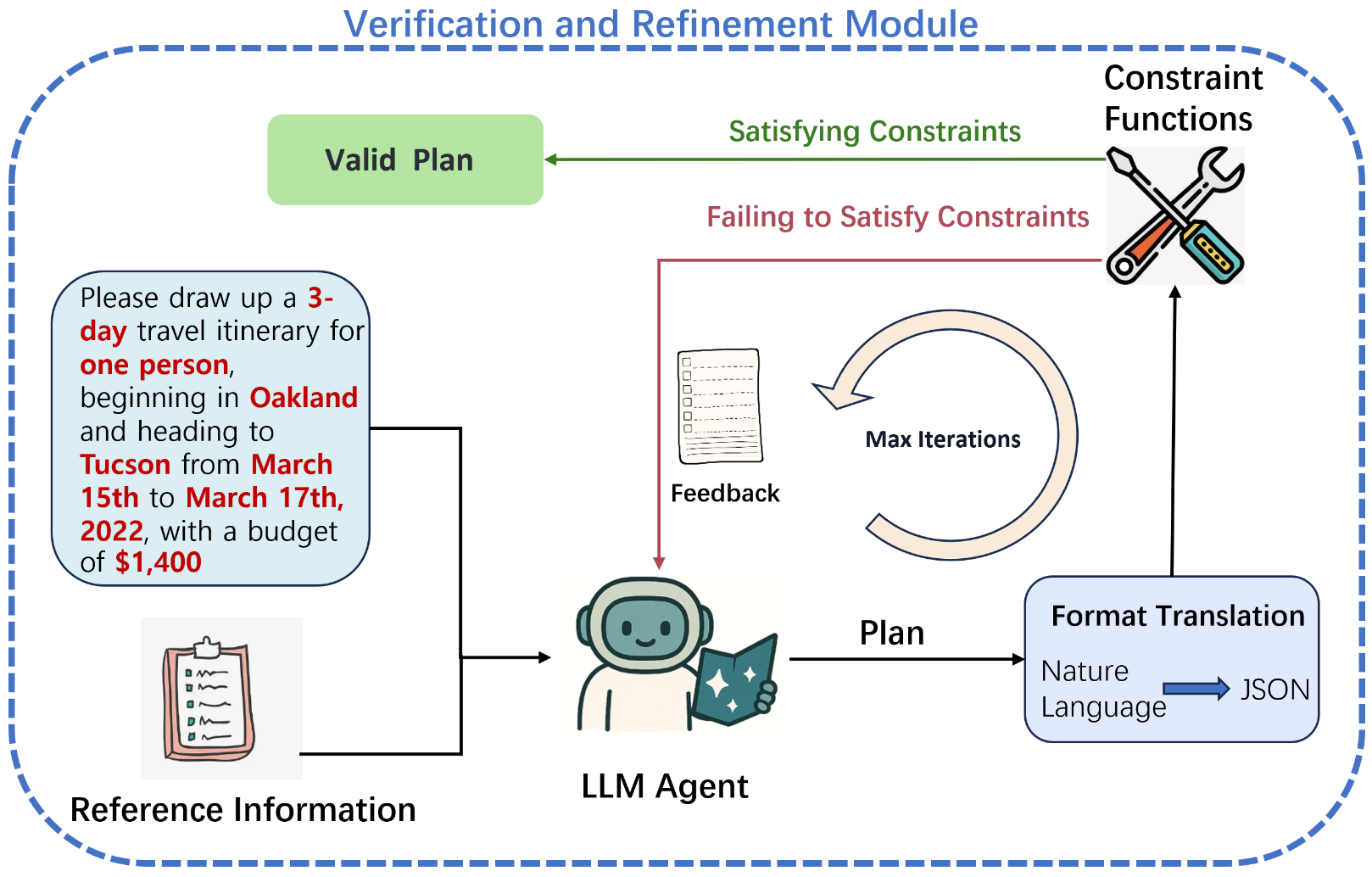}
    \caption{An overview of the Verification and Refinement Module, which illustrates the iterative process where the LLM agent generates a plan, verifies it against constraints, and refines it based on feedback.}
    \label{fig:verification_refinement} 
\end{figure}

\subsection{Incremental Merge}
\label{sec:merge}
Once the local agents complete the generation and validation of subtask solutions, the system proceeds to the incremental merging stage, which is responsible for integrating the subplans into a unified global plan. 

Incremental merging does not directly combine all subplans, but follows an order from simple to complex. 
In the task of travel planning, it first combines the transportation and attraction subplans to establish a basic itinerary, then incorporates accommodation arrangements, and finally completes the plan based on user meals preferences. Each phase of the integration relies on the results of the previous phase, ensuring contextual coherence throughout the plan. 
The specific merging operations are executed by an LLM agent equipped with a prompt template (provided in Appendix \ref{prompt:merge}) and constraint evaluation function. 
Since the local agent generate multiple subplans for each subtask, we employ the Cartesian product to combine all subplans generated from each local agent, and attempt to merge subplans for all possible combinations in pursuit of the optimal solution. 
If a candidate solution passes all constraint checks, it is selected as the final plan. If no solution fully satisfies all constraints, the one that meets the most constraints is chosen as the suboptimal alternative.


\subsection{Verification and Refinement Module}
\label{sec:verfify}
Although complex tasks are decomposed into multiple more manageable subtasks, local agents may still generate subplans that violate local constraints such as mismatched transportation types. Furthermore, the merging process can also produce failed plans that breach global constraints such as exceeding the total budget. 
Inspired by the iterative self-feedback mechanism ~\cite{madaan2023self, kambhampati2024llmmodulo}, we propose Verification and Refinement Module that checks whether plans satisfy the constraints and guides agents in iteratively improving their plans when they fail to meet these constraints. This module is applied in the Local Plan Generation and Incremental Merge stages. 

Figure~\ref{fig:verification_refinement} illustrates the Verification and Refinement Module. 
Once any agent generates a formatted plan, constraint evaluation functions are employed to verify its constraint satisfaction. When constraints are violated, the generated plan along with the constraint check results are fed back to the LLM agent for re-planning, using the prompt provided in Appendix \ref{prompt:verification}. 
The re-planning process terminates until constraints are satisfied or the maximum iteration limit is reached.


\section{Experiment}
\begin{table*}[!t]
\centering
\caption{Results of different methods on the TravelPlanner validation set. The best results are marked in \textbf{bold}. }
\label{table:main}
\resizebox{0.85\textwidth}{!}{%
\begin{tabular}{lcccccc}
\toprule
\multirow{2}{*}{Method} & Delivery & \multicolumn{2}{c}{Commonsense} & \multicolumn{2}{c}{Hard Constraint} & Final \\
\cmidrule(lr){3-4} \cmidrule(lr){5-6}
& Rate & Micro & Macro & Micro & Macro & Pass Rate \\
\midrule

\multicolumn{7}{c}{\textbf{Qwen2.5-32B-Instruct}} \\

Direct \cite{xie2024travelplanner} & 100 & 79.5 & 16.1 & 30.0 & 16.7 & 2.8 \\
CoT \cite{wei2022chain} & 100 & 76.9 & 11.7 & 25.7 & 13.3 & 0.6 \\
LLM-Modulo \cite{gundawar2024modulotravel} & 100 & 90.4 & 46.7 & 71.9 & 59.4 & 36.7 \\
DPPM & \textbf{100} & \textbf{95.3} & \textbf{70.0} & \textbf{80.9} & \textbf{69.4} & \textbf{58.9} \\
\midrule

\multicolumn{7}{c}{\textbf{Qwen2.5-72B-Instruct}} \\

Direct \cite{xie2024travelplanner} & 100 & 83.7 & 21.1 & 34.5 & 18.3 & 6.6 \\
CoT \cite{wei2022chain} & 100 & 82.8 & 21.1 & 41.0 & 22.2 & 6.6 \\
LLM-Modulo \cite{gundawar2024modulotravel} & 100 & 92.6 & 53.3 & 67.6 & 52.2 & 39.4 \\
DPPM & \textbf{100} & \textbf{96.9} & \textbf{77.8} & \textbf{82.6} & \textbf{73.3} & \textbf{64.4} \\
\midrule

\multicolumn{7}{c}{\textbf{DeepSeek-V3}} \\

Direct \cite{xie2024travelplanner} & 100 & 81.4 & 25.6 & 39.5 & 20.0 & 6.7 \\
CoT \cite{wei2022chain} & 100 & 84.2 & 23.9 & 40.9 & 21.1 & 7.8 \\
LLM-Modulo \cite{gundawar2024modulotravel} & 100 & 94.8 & 69.4 & 86.7 & 76.7 & 57.8 \\
DPPM & \textbf{100} & \textbf{98.5} & \textbf{89.4} & \textbf{90.7} & \textbf{80.6} & \textbf{76.7} \\
\midrule

\multicolumn{7}{c}{\textbf{DeepSeek-R1}} \\

Direct \cite{xie2024travelplanner} & 100 & 84.0 & 27.8 & 59.5 & 47.8 & 17.8 \\
CoT \cite{wei2022chain} & 100 & 85.9 & 37.8 & 50.2 & 41.7 & 25.0 \\
DPPM & \textbf{100} & \textbf{98.4} & \textbf{90.6} & \textbf{94.5} & \textbf{90.6} & \textbf{87.2} \\


\bottomrule
\end{tabular}%
}
\end{table*}

\begin{table*}[!t]
\centering
\vspace{10pt}
\caption{Results of different methods on ChinaTravel dataset across three difficulty levels on Qwen2.5-32B-Instruct backbone. The best results are marked in \textbf{bold}. }
\label{table:main2}
\resizebox{1.0\textwidth}{!}{%
\small 
\begin{tabular}{llcccccc}
\toprule
Difficulty & Method & Delivery & \multicolumn{2}{c}{Commonsense} & \multicolumn{2}{c}{Hard Constraint} & Final \\
\cmidrule(lr){4-5} \cmidrule(lr){6-7}
 &  & Rate & Micro & Macro & Micro & Macro & Pass Rate \\
\midrule

\multirow{3}{*}{Easy} 
 & Direct \cite{xie2024travelplanner} & 100 & 82.8 & 22.5 & 95.7 & 75.0 & 16.2 \\
 & LLM-Modulo \cite{gundawar2024modulotravel} & 100 & 94.4 & 73.8 & 96.4 & 78.3 & 55.8 \\
 & DPPM & \textbf{100} & \textbf{98.6} & \textbf{92.5} & \textbf{98.8} & \textbf{92.5} & \textbf{87.1} \\
\cmidrule(r){1-8}

\multirow{3}{*}{Medium} 
 & Direct \cite{xie2024travelplanner} & 100 & 80.0 & 14.0 & 87.6 & 43.3 & 6.0 \\
 & LLM-Modulo \cite{gundawar2024modulotravel} & 100 & 92.3 & 61.3 & 93.2 & 65.3 & 38.7 \\
 & DPPM & \textbf{100} & \textbf{95.3} & \textbf{78.7} & \textbf{97.1} & \textbf{84.0} & \textbf{70.7} \\
\cmidrule(r){1-8}

\multirow{3}{*}{Human} 
 & Direct \cite{xie2024travelplanner} & 100 & 84.9 & 19.1 & 88.5 & 48.9 & 9.9 \\
 & LLM-Modulo \cite{gundawar2024modulotravel} & 100 & 95.9 & 72.3 & 90.3 & 52.5 & 39.7 \\
 & DPPM & \textbf{100} & \textbf{97.0} & \textbf{76.6} & \textbf{93.1} & \textbf{63.8} & \textbf{52.5} \\

\bottomrule
\end{tabular}%
}
\end{table*}
\subsection{Experimental Setup}
To evaluate the performance of DPPM in real-world multi-constraints planning scenarios, we conducted experiments on the TravelPlanner \cite{xie2024travelplanner} dataset and a modified version of the ChinaTravel \cite{shao2024chinatravel} dataset.
In all experiments, we follow the \textit{sole-planning} setting in TavelPlanner ~\cite{xie2024travelplanner}, where LLMs generate plans based on the provided necessary information without requiring tool use such as searching the database for the departure time of a flight. 

\textbf{TravelPlanner} \cite{xie2024travelplanner}. 
In TravelPlanner, given user-specified origin, destination, personal requirements, and relevant reference information, LLM agents were required to generate a travel plan that satisfies various constraints. 
The dataset defines 13 constraints, categorized into two types: commonsense constraints and hard constraints. 
Among these, 5 constraints are global and 8 constraints are local. 
Details of these constraints can be found in Appendix \ref{appendix:definition}. 
The travel plans generated by LLM must include detailed information across these dimensions: day, current\_city, transportation, breakfast, attraction, lunch, dinner, and accommodation.

\textbf{ChinaTravel-M} \cite{shao2024chinatravel}
is a benchmark specifically designed for authentic Chinese travel planning scenarios, requiring generated plans to include arrangements for attractions, restaurants, accommodations, and transportation between events.
We adapt the original ChinaTravel dataset by modifying its output format and evaluation process to align with the TravelPlanner dataset, which we name ChinaTravel-M. 
More details about these adaptations are provided in Appendix \ref{appendix:dataset}.

\textbf{Metrics.}
For both datasets, we evaluate the plans using four metrics: 
(1) \textbf{Delivery Rate.} The ratio of samples in which the agent can successfully generate a plan within the limited number of steps. (2) \textbf{Commonsense Constraint Pass Rate.}  The ratio of commonsense constraints that are satisfied. (3) \textbf{Hard Constraint Pass Rate.} The ratio of hard constraints that are satisfied. (4) \textbf{Final Pass Rate.} The ratio of plans satisfying all commonsense and hard constraints, and it is the most critical metric.

We evaluate the Commonsense Constraint Pass Rate and Hard Constraint Pass Rate using \textit{micro} and \textit{macro} metrics, where \textit{micro} measures the proportion of passed constraints out of all constraints, while \textit{macro} measures the proportion of fully compliant plans among all tested plans.

\textbf{Baselines.}
We compare our method with the following sole-planning baselines:
(1) \textbf{Direct} \cite{xie2024travelplanner} utilizes LLMs to directly generate the travel plan based on the provided reference information.
(2) \textbf{CoT} \cite{wei2022chain} employs chain-of-thought instructions to guide LLMs in generating intermediate reasoning steps, thus activating their reasoning capabilities for systematic task decomposition and planning.
(3) \textbf{LLM-Modulo} \cite{kambhampati2024llmmodulo, gundawar2024modulotravel} integrates LLM with reliable constraint evaluation functions in a generate-test framework, utilizing verification results as feedback to the LLM for solving challenging planning problems. 
The description of base models in all methods can be found in Appendix \ref{appendix:param}.

\subsection{Implementation Details}
To keep prompts consistent in style and avoid over-designing,  we use either the original prompt templates from TravelPlanner \cite{xie2024travelplanner} or modified versions based on the original prompt templates for all methods (including ours).
Specifically, to support local plan generation or plan refinement, the prompt templates in DPPM and LLM-Modulo require corresponding modifications.
More details about prompt templates for our method are available in Appendix \ref{appendix:prompt}.

The reference information provided in both the TravelPlanner and ChinaTravel-M datasets contains redundant details that could lead to LLM hallucinations due to the lengthy context, rather than focusing on planning. 
To better evaluate the planning capabilities of different methods, we removed redundant reference information from both datasets for all methods.
Specifically, for the TravelPlanner dataset, we removed phone numbers, website links, and specific geocoordinates from the reference information. 
For the ChinaTravel-M dataset, we removed specific geocoordinates from the reference information. 
In our experiments, we set the maximum number of iterations for plan refinement using outputs from constraint evaluation functions to $10$.

Furthermore, our method employs the same constraint evaluation functions and verification result formats as LLM-Modulo \cite{gundawar2024modulotravel}.
The implementations of constraint evaluation functions are adapted from the evaluation code of the TravelPlanner and ChinaTravel-M datasets.

\begin{figure}[!t]
  \centering
    \centering
    \includegraphics[width=1\linewidth]{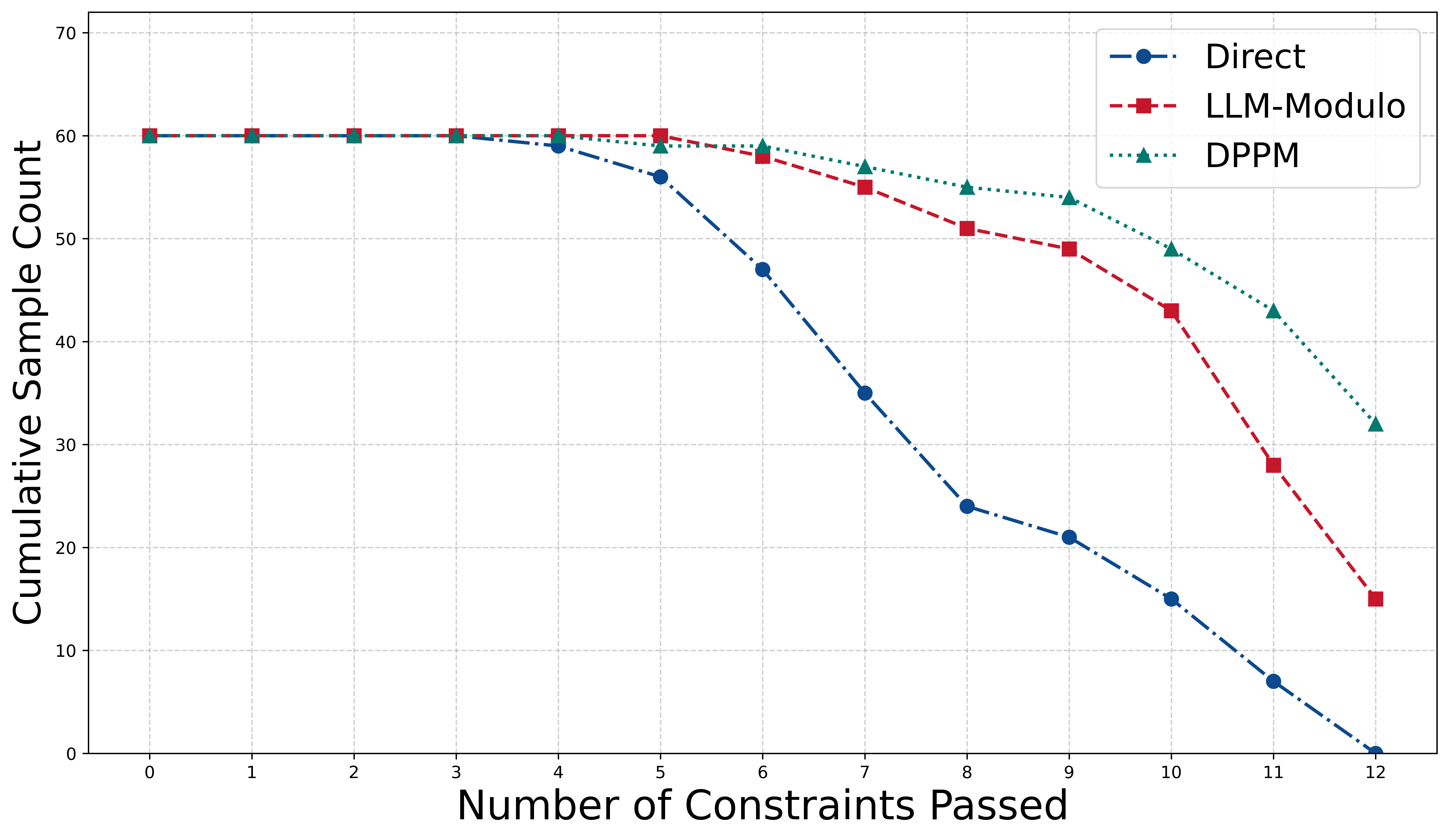}
    \caption{
    The number of passed samples across the number of constraints for baselines and our method on Hard-level samples of TravelPlanner. 
    }
    \label{fig:constraint} 
\end{figure}

\begin{figure}[!t]
  \centering
    \centering
    \includegraphics[width=1\linewidth]{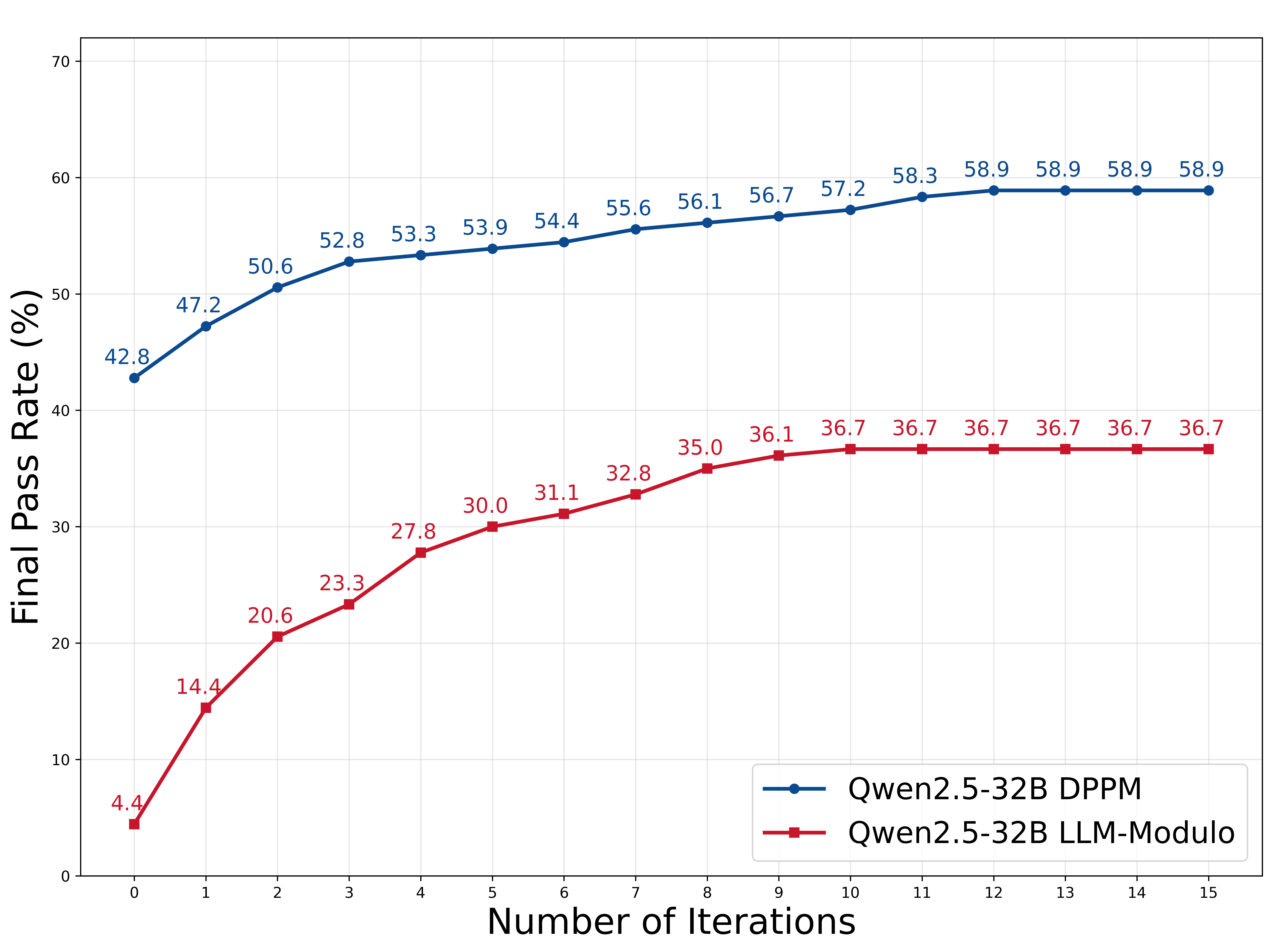}
    \caption{
    Comparison of Final Pass Rate between baselines and our method across the number of self-refinement iterations on TravelPlanner dataset. The base model is Qwen2.5-32B-Instruct.
    } 
    \label{fig:iter} 
\end{figure}

\subsection{Main Results}
In Table \ref{table:main}, the proposed DPPM can consistently outperform baselines in the TavelPlanner dataset.
Compared to baselines, our model achieves significant improvements in Commonsense, Hard Constraint, and the overall Final Pass Rate.
Specifically, when using Qwen2.5-32B-Instruct as the base model, our method outperforms Direct by 56.1\%, CoT by 58.3\%, and LLM-Modulo by 22.2\% on final pass rate.
With Qwen2.5-72B-Instruct, DeepSeek-V3, and DeepSeek-R1 as base models, our method outperforms Direct on Final Pass Rate by 57.8\%, 70.0\%, and 69.4\% respectively. 
Although CoT decomposes planning tasks step-by-step through sequential reasoning, it is prone to hallucinations and cascading errors, resulting in a lower Final Pass Rate than Direct when using Qwen2.5-32B-Instruct as the base model.
Both LLM-Modulo and DPPM utilize the same constraint evaluation functions for plan self-refinement, but DPPM consistently outperforms LLM-Modulo across all metrics, demonstrating the superior effectiveness of our method.

\begin{table*}[!th]
\centering
\caption{Results of Qwen2.5-32B-Instruct on different ablation study settings. The best results are marked in \textbf{bold}. The symbol '-' denotes the removal of the component. }
\label{table:ablation}
\resizebox{0.95\textwidth}{!}{%
\begin{tabular}{lcccccc}
\toprule

\multirow{2}{*}{Method} & Delivery & \multicolumn{2}{c}{Commonsense Pass Rate} & \multicolumn{2}{c}{Hard Constraint Pass Rate} & Final \\
\cmidrule(lr){3-4} \cmidrule(lr){5-6}
& Rate & Micro & Macro & Micro & Macro & Pass Rate \\
\midrule


DPPM & \textbf{100} & \textbf{95.3} & \textbf{70.0} & 80.9 & \textbf{69.4} & \textbf{58.9} \\
- Global Constraint Instruction & 100 & 93.4 & 60.7 & \textbf{81.5} & 62.3 & 49.4 \\
- Verification and Refinement & 100 & 81.9 & 25.0 & 49.8 &  43.9 &  20.0 \\
- Cartesian Product & 100 & 94.4 & 64.4 & 79.5 & 60.8 & 46.1 \\

\bottomrule
\end{tabular}%
}
\end{table*}

As shown in Table \ref{table:main2}, we observe consistent results in which our method significantly outperforms the baselines on all difficulty levels in the ChinaTravel-M dataset.
Using the smaller Qwen2.5-32B-Instruct as the base model, our method still performs excellently on all metrics.
Specifically, on Easy-level samples, our method achieves 70.9\% and 31.3\% higher Final Pass Rate than Direct and LLM-Modulo, respectively.
Our method outperforms LLM-Modulo on Final Pass Rate by 32.0\% on Medium-level samples and 12.8\% on Human-level samples, respectively.

Notably, among the four metrics, our method demonstrates the most significant improvements in macro Commonsense and macro Hard Constraint Pass Rate on both the TravelPlanner and ChinaTravel datasets.
On both datasets, while Direct achieves reasonably high micro Commonsense / Hard Constraint Pass Rate, its macro Commonsense / Hard Constraint Pass Rate remains substantially lower than the micro-level counterparts.
This demonstrates that while plans directly generated by LLMs can adequately satisfy a few constraints, they are consistently hard to meet all constraints simultaneously.

\textbf{Analysis on the Satisfying Constraints.} As shown in Figure \ref{fig:constraint}, for Hard-level samples containing 12 constraints, the number of passed samples in both Direct and LLM-Modulo decreases rapidly with increasing constraint counts, whereas DPPM shows a slower decline rate.
Specifically, most Direct-generated plans satisfy 5 constraints, but beyond this threshold, the number of passed plans decrease sharply in an approximately linear fashion.
DPPM and LLM-Modulo show no significant drop in the number of passed plans until exceeding 10 constraints, but DPPM achieves double the number of plans that satisfy all constraints compared to LLM-Modulo.
This demonstrates that our method's plan decomposition and merging operations effectively enhance the model's capability to handle planning tasks with complex constraints.

\textbf{Effectiveness of the Verification and Refinement Module.} Figure \ref{fig:iter} shows how DPPM and LLM-Modulo's Final Pass Rate changes with the number of verification-refinement iterations. 
As shown in Figure \ref{fig:iter}, as the number of iterations increases, the Final Pass Rate converges to its maximum value.
When agents cannot improve plans with correct verification feedback, it indicates the method has reached its performance limit.
DPPM achieves a higher Final Pass Rate ceiling than LLM-Modulo, showing better performance and potential.

\begin{figure}[!t]
  \centering
    \centering
    \vspace{10pt}
    \includegraphics[width=1\linewidth]{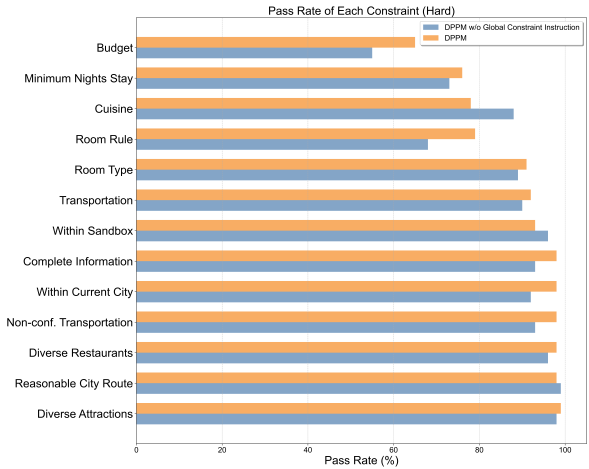}
    \caption{Comparison of Pass Rate for each constraint between with and without Global Constraint Instruction on Hard-level samples of TravelPlanner dataset. }
    \label{fig:global_constraint} 
\end{figure}

\subsection{Ablation Study}

To evaluate the impact of the global constraint instruction and the verification and refinement module, we conduct ablation studies using Qwen2.5-32B-Instruct as the base model on the TravelPlanner dataset.

\textbf{Global Constraint Instruction.} As shown in Table \ref{table:ablation}, without the global constraint instruction in prompt templates, the performance of our method drops by 9.3\% in macro Commonsense Pass Rate, 7.1\% in Hard Constraint Pass Rate, and 9.5\% in Final Pass Rate.
It shows the effectiveness of the global constraint instruction.
As shown in Figure \ref{fig:global_constraint}, without the global constraint instruction, our method shows significant drops in pass rates for global constraints, specifically the Budget constraint and Minimum Nights Stay constraint.

\textbf{Verification and Refinement.} We also consider the impact of the verification and refinement module of our method.
As shown in Table \ref{table:ablation}, when we completely remove the verification and refinement module from our method, the Final Pass Rate drops from 58.9\% to 20.0\%, which shows this module's important role in enhancing the agent's planning capability.
However, as shown in Table \ref{table:ablation} and Table \ref{table:main}, even without the verification and refinement module, our method still significantly outperforms Direct and CoT.
Among methods without external verification, ours achieves state-of-the-art performance.

\textbf{Cartesian Product.} To evaluate the impact of the Cartesian Product combination in the Incremental Merge module in Section \ref{sec:merge}, we compare the performance with and without Cartesian Product on the TravelPlanner dataset.
When the Cartesian Product combination is disabled, we combine only the first plan from each local agent to generate a candidate solution.
As shown in Table \ref{table:ablation}, without the Cartesian Product combination, our method shows a 12.8\% drop in Final Pass Rate performance.

\section{Conclusion}
In this work, we propose DPPM, a novel paradigm for LLM-based multi-constraints planning. 
The core idea of DPPM is to decompose complex tasks into subtasks based on constraints, generate subplans for each subtask in parallel, and merge them into a coherent global plan. 
This constraint-aware decomposition significantly reduces the burden on LLM, leading to substantial performance improvements. 
To address potential errors introduced during local planning and the merging stage, we incorporate a verification and refinement module that systematically corrects errors and resolves conflicts. 
Experimental results on two travel planning benchmarks demonstrate that our method significantly outperforms existing baselines and validate the effectiveness of each component in our paradigm. 

\section*{Limitations}
Our work represents an significant step forward for enhancing LLM-based agents' planning abilities, but it is not without limitations.
One limitation lies in its adaptability to diverse types of planning problems. 
DPPM focuses on solving planning tasks with complex constraints.
Since DPPM relies on constraint-based task decomposition before planning, it cannot fully leverage its strengths when handling single-constraint planning tasks.
Future work will focus on extending our method to more general forms of planning tasks.

\bibliography{reference}
\clearpage
\newpage
\appendix

\lstset{
    backgroundcolor=\color{gray!5},
    basicstyle=\small\ttfamily\linespread{1.1},
    breaklines=true,
    frame=lines,
    rulecolor=\color{gray!30},
    emph={Example,Query,Travel Plan,Day,Current City,Transportation,
          Breakfast,Lunch,Dinner,Accommodation,Attraction},
}

\section{ChinaTravel-M Dataset}
\label{appendix:dataset}


The ChinaTravel dataset encodes travel planning rules through string-formatted Constraint Expressions (CEs).
To adapt the original ChinaTravel dataset by modifying its output format and evaluation process to align with the TravelPlanner dataset, we performed the following data processing on the ChinaTravel dataset:

(1) Semantic Parsing: Raw CEs are converted into structured constraints covering budget ranges, POI categories, accommodation standards, and dining preferences through contextual understanding and logical decomposition ;

(2) Iterative Refinement: Model-guided multi-round adjustments ensure logical accuracy and format compliance with TravelPlanner’s requirements;

(3) Standardized Mapping: Parsed constraints are transformed into numerical parameters (e.g., budget limits) and formalized representations via schema-driven generation.


\section{Example of Prompts}
\label{appendix:prompt}
\subsection{Verification and Refinement PROMPT}
\label{prompt:verification}
We provide the instruction prompt for Verification and Refinement as follows, which is designed to guide the model in iteratively improving travel plan based on feedback.

\begin{lstlisting}
You are a proficient planner. Based on the provided information and query, please give me a detailed plan, including specifics such as flight numbers (e.g., F0123456), restaurant names, and accommodation names. Note that all the information in your plan should be derived from the provided data. You must adhere to the format given in the example. Additionally, all details should align with commonsense. The symbol '-' indicates that information is unnecessary. For example, in the provided sample, you do not need to plan after returning to the departure city. When you travel to two cities in one day, you should note it in the 'Current City' section as in the example (i.e., from A to B).
Your task is to revise the plan according to feedback from Critics. All the information needed for generating the plan can be obtained from the Given information. 

Important Note:
- Do NOT include any comments or explanations in the JSON output
- Each field should only contain its value without any additional notes or comments
- The output should be pure JSON without any annotations or explanations

Query:
{query}

Given Information:
<REFERENCE_INFORMATION_START>
{reference_information}
<REFERENCE_INFORMATION_END>

Current Plan:
{current_plan}

Feedback from Critics:
<FEEDBACK_START>
{feedback}
<FEEDBACK_END>

Output Format
Please provide the revised travel plan in the following JSON format:

{{
  "plan": [
    {{
        "days": <int>,  # Day number
        "current_city": <str>,  # Current city or travel route
        "transportation": <str>,  # Transportation details (or "-" if not nessesary)
        "breakfast": <str>,  # Breakfast details (or "-" if not nessesary)
        "attraction": <str>,  # Attraction details (or "-" if not nessesary)
        "lunch": <str>,  # Lunch details (or "-" if not nessesary)
        "dinner": <str>,  # Dinner details (or "-" if not nessesary)
        "accommodation": <str>  # Accommodation details (or "-" if not nessesary)
    }},
    ...
  ]
}}
Do not output any other content, please ensure that the output content can be parsed by json.loads()
Ensure the output strictly adheres to this format.

Travel Plan in JSON format:
\end{lstlisting}

\subsection{Local Plan Generation PROMPT}
\label{prompt:loacl_plan}
We provide the instruction prompt for Local Plan Generation as follows, which is designed to guide the model in generating local plan.

\begin{lstlisting}
You are a proficient planner. Based on the provided information and query, please give me multiple detailed accommodation plans (at least 2 different plans), including specifics such as accommodation names. Note that all the information in your plans should be derived from the provided data. Additionally, all details should align with commonsense. The symbol "-" indicates that information is unnecessary. When you travel to two cities in one day, you should note it in the "Current City" section as in the example (i.e., from A to B).
You must generate the plan solely based on the information provided in the reference, without using any of your internal information. In the description, you need to explain why you chose to generate the plan in that manner and whether your plan meets the requirements of the query.
You must adhere to the format given in the example. In particular, in every plan, the value for "accommodation" must include both the accommodation name and its corresponding city, separated by a comma (for example: "Affordable Spacious Refurbished Room in Bushwick!, Charlotte").
In the following examples, the parts marked as <omit> must not be omitted when you actually generate the plan; you need to provide the specific details.
Provide your multiple plans in JSON structure like this example:

***** Example Starts *****
Query: Could you create a travel plan for 7 people from Ithaca to Charlotte spanning 3 days, from March 8th to March 14th, 2022, with a budget of $30,200?
Travel Plan in JSON format:
[
  {{
    "plan_id": 1,
    "description": "Accommodation Plan 1",
    "plan": [
      {{
        "days": 1,
        "current_city": "from Ithaca to Charlotte",
        "accommodation": "Affordable Spacious Refurbished Room in Bushwick!, Charlotte"
      }},
      {{
        "days": 2,
        "current_city": "Charlotte",
        "accommodation": "Affordable Spacious Refurbished Room in Bushwick!, Charlotte"
      }},
      {{
        "days": 3,
        "current_city": "from Charlotte to Ithaca",
        "accommodation": "-"
      }}
    ]

  }},
  {{
    "plan_id": 2,
    "description": "",
    "plan": [
      {{
        "days": 1,
        "current_city": "",
        "accommodation": ""
      }},
      {{
        "days": 2,
        "current_city": "",
        "accommodation": ""
      }},
      {{
        "days": 3,
        "current_city": "",
        "accommodation": ""
      }}
    ]

  }}
]
***** Example Ends *****

Important Instructions:
1. You must strictly adhere to the total budget specified in the query.
2. This is a local constraint dimension (e.g., accommodations). You should allocate the minimum necessary budget to this dimension while leaving as much budget redundancy as possible for other dimensions (e.g., transportation, attractions, meals).
3. Prioritize selecting lower-cost accommodations when multiple choices are available.
4. Ensure that the cumulative cost of all accommodations does not exceed the allocated budget for this dimension.

Given reference information: 
<REFERENCE_INFORMATION_START>
{text}
<REFERENCE_INFORMATION_END>

Query: 
{query}

Do not output any other content, please ensure that the output content can be parsed by json.loads()
Travel Plan in JSON format:
\end{lstlisting}

\subsection{Incremental Merge PROMPT}
\label{prompt:merge}
We provide the instruction prompt for Incremental Merge as follows, which is designed to guide the model in incrementally integrating local plans into a unified global plan.

\begin{lstlisting}
You are a proficient planner. Based on the provided information and query, please give me a detailed plan, including specifics such as flight numbers (e.g., F0123456), restaurant names, and accommodation names. Note that all the information in your plan should be derived from the provided data. You must adhere to the format given in the example. Additionally, all details should align with commonsense. The symbol "-" indicates that information is unnecessary. For example, in the provided sample, you do not need to plan after returning to the departure city. When you travel to two cities in one day, you should note it in the "Current City" section as in the example (i.e., from A to B).

Your task is to merge and revise two flawed partial plans based on feedback from critics into one that meets the requirements. All the information needed for generating the plan can be obtained from the Given Information. 

Important Note:
- Do NOT include any comments or explanations in the JSON output
- Each field should only contain its value without any additional notes or comments
- The output should be pure JSON without any annotations or explanations
- If transportation-related arrangements exist,please note that self-driving and flight transportation methods are conflicting and cannot appear in the same plan.For example, if flight transportation is scheduled on either the first or the last day, then self-driving arrangements must not appear on any of the remaining days.

***** Example Starts *****
Query: Could you create a travel plan for 7 people from Ithaca to Charlotte spanning 3 days, from March 8th to March 14th, 2022, with a budget of $30,200?

Travel Plan in JSON format:
{example_plan}
***** Example Ends *****

Query:
{query}

Given Information:
<REFERENCE_INFORMATION_START>
{reference_information}
<REFERENCE_INFORMATION_END>

Partial Plan A:
{plan_a}

Feedback for Plan A:
<FEEDBACK_START_A>
{feedback_a}
<FEEDBACK_END_A>

Partial Plan B:
{plan_b}

Feedback for Plan B:
<FEEDBACK_START_B>
{feedback_b}
<FEEDBACK_END_B>

Merged Correct Plan:
\end{lstlisting}

\section{Constraints Definition}
We divide all the constraints into two categories: local constraints and global constraints. The specific definitions are shown in Figure~\ref{fig:constraint}.
\label{appendix:definition}
\begin{figure}[!h]
  \centering
    \centering
    \includegraphics[width=1\linewidth]{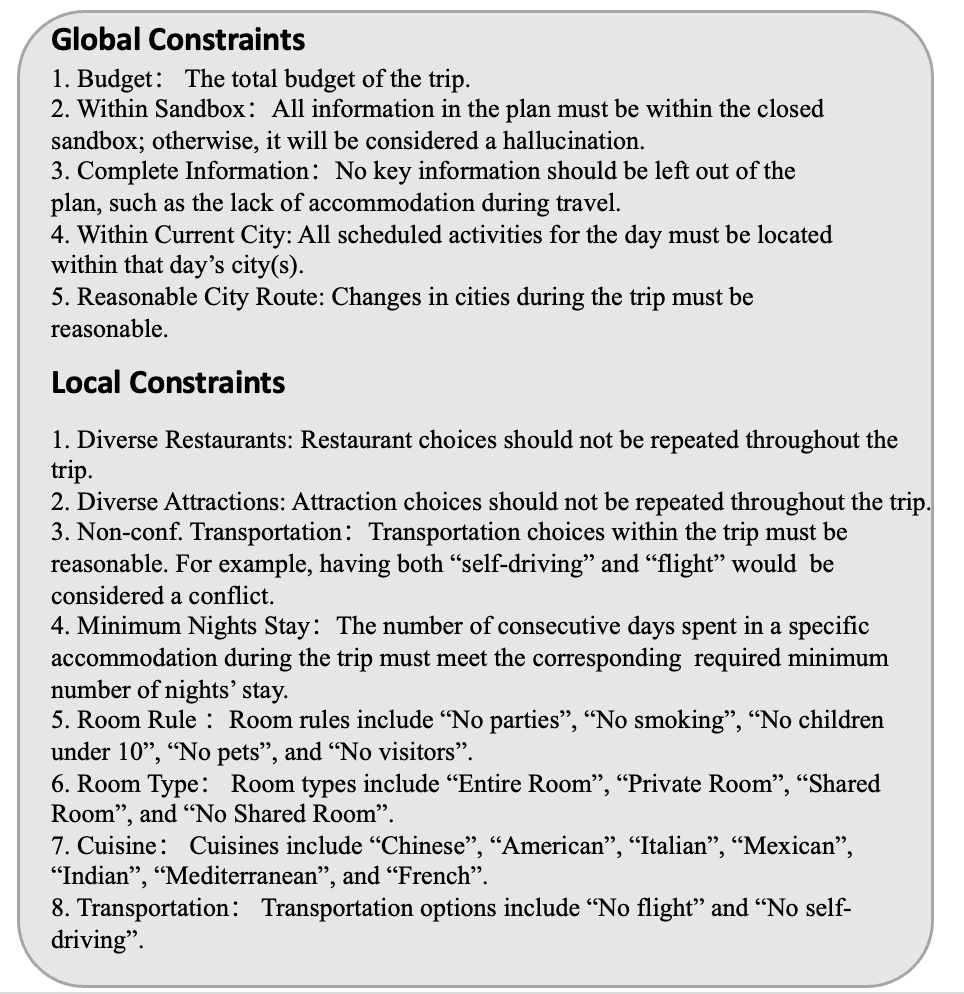}
    \caption{Constraints Definition.}
    \label{fig:constraints} 
\end{figure}

\section{Number of Parameters in Base Models}
\label{appendix:param}

The number of parameters in each base model is shown in Table \ref{tab:param}.

\begin{table}[!h]
	\centering
	\scalebox{0.88}{
		\begin{tabular}{l| c}
			\toprule                  
			 {Model} & {Number of Parameters (B)} \\		
			\midrule 
                Qwen2.5-32B-Instruct & 32 \\
                Qwen2.5-72B-Instruct & 72 \\
                Deepseek-V3 & 671 \\
                Deepseek-R1 & 671 \\
			\bottomrule
	\end{tabular}}
	\caption{Number of parameters in base models.
    }
    \label{tab:param}
\end{table}

\end{document}